\documentclass[runningheads]{llncs}
\usepackage{graphicx}
\usepackage{booktabs}
\usepackage{amssymb}
\usepackage{hyperref}
\begin{document}
%
\title{A Multi-cascaded Deep Model for Bilingual SMS Classification}
%
%
\author{Muhammad Haroon Shakeel\orcidID{0000-0001-6237-3388} \and
Asim Karim\orcidID{0000-0002-9872-5020} \and
Imdadullah Khan\orcidID{0000-0002-6955-6168}}

\authorrunning{M. H. Shakeel et al.}
%
\institute{Department of Computer Science, Syed Babar Ali School of Science and Engineering, Lahore University of Management Sciences (LUMS), Lahore, Pakistan
\email{\{15030040,akarim,imdad.khan\}@lums.edu.pk}}
\maketitle              
\begin{abstract}
Most studies on text classification are focused on the English language. However, short texts such as SMS are influenced by regional languages. This makes the automatic text classification task challenging due to the multilingual, informal, and noisy nature of language in the text. In this work, we propose a novel multi-cascaded deep learning model called \emph{McM} for bilingual SMS classification. McM exploits $n$-gram level information as well as long-term dependencies of text for learning. Our approach aims to learn a model without any code-switching indication, lexical normalization, language translation, or language transliteration. The model relies entirely upon the text as no external knowledge base is utilized for learning. For this purpose, a $12$ class bilingual text dataset is developed from SMS feedbacks of citizens on public services containing mixed Roman Urdu and English languages. Our model achieves high accuracy for classification on this dataset and outperforms the previous model for multilingual text classification, highlighting language independence of McM.

\end{abstract}
\section{Introduction} \label{sec:introduction}
Social media such as Facebook, Twitter, and Short Text Messaging Service (SMS) are popular channels for getting feedback from consumers on products and services. In Pakistan, with the emergence of e-government practices, SMS is being used for getting feedback from the citizens on different public services with the aim to reduce petty corruption and deficient delivery in services. Automatic classification of these SMS into predefined categories can greatly decrease the response time on complaints and consequently improve the public services rendered to the citizens. While Urdu is the national language of Pakistan, English is treated as the official language of the country. This leads to the development of a distinct dialect of communication known as Roman Urdu, which utilizes English alphabets to write Urdu. Hence, the SMS texts contain multilingual text written in the non-native script and informal diction. The utilization of two or more languages simultaneously is known as multilingualism~\cite{fatima2018multilingual}. Consequently, alternation of two languages in a single conversation, a phenomenon known as code-switching, is inevitable for a multilingual speaker~\cite{williams2019bilinguals}. Factors like informal verbiage, improper grammar, variation in spellings, code-switching, and short text length make the problem of automatic bilingual SMS classification highly challenging. 

In Natural Language Processing (NLP), deep learning has revolutionized the modeling and understanding of human languages. The richness, expressiveness, ambiguities, and complexity of the natural language can be addressed by deep neural networks without the need to produce complex engineered features~\cite{denecke2008using}. Deep learning models have been successfully used in many NLP tasks involving multilingual text. A Convolutional Neural Network (CNN) based model for sentiment classification of a multilingual dataset was proposed in~\cite{medrouk2017deep}. However, a particular record in the dataset belonged to one language only. In our case, a record can have either one or two languages. There is very little published work on this specific setting. One way to classify bilingual text is to normalize the different variations of a word to a standard spelling before training the model~\cite{rafae2015unsupervised}. However, such normalization requires external resources such as lexical database, and Roman Urdu is under-resourced in this context. Another approach for an under-resourced language is to adapt the resources from resource-rich language~\cite{zhou2016attention}. However, such an approach is not generalizable in the case of Roman Urdu text as it is an informal language with no proper grammatical rules and dictionary. More recent approach utilizes code-switching annotations to improve the predictive performance of the model, where each word is annotated with its respective language label. Such an approach is not scalable for large data as annotation task becomes tedious. 
 
 In this paper, we propose a multi-cascaded deep learning network, called as \emph{McM} for multi-class classification of bilingual short text. Our goal is to achieve this without any prior knowledge of the language, code-switching indication, language translation, normalizing lexical variations, or language transliteration. In multilingual text classification, previous approaches employ a single deep learning architecture, such as CNN or Long Short Term Memory (LSTM) for feature learning and classification. McM, on the other hand, employs three cascades (aka feature learners) to learn rich textual representations from three perspectives. These representations are then forwarded to a small discriminator network for final prediction.  We compare the performance of the proposed model with existing CNN-based model for multilingual text classification~\cite{medrouk2017deep}. We report a series of experiments using $3$ kinds of embedding initialization approaches as well as the effect of attention mechanism~\cite{wang-etal-2016-bilingual}.

The English language is well studied under the umbrella of NLP, hence many resources and datasets for the different problems are available. However, research on English-Roman Urdu bilingual text lags behind because of non-availability of gold standard datasets. Our second contribution is that we present a large scale annotated dataset in Roman Urdu and English language with code-switching, for multi-class classification. The dataset consists of more than $0.3$ million records and has been made available for future research.

The rest of the paper is organized as follows. Section~\ref{sec:datasetDescription} defines the dataset acquiring process and provides an explanation of the class labels. In section~\ref{sec:proposedModel}, the architecture of the proposed model, its hyperparameters, and the experimental setup is discussed. We discuss the results in section~\ref{sec:results} and finally, concluding remarks are presented in section~\ref{sec:conclusion}.

 .    

\section{Dataset Acquisition and Description} \label{sec:datasetDescription}

\begin{table}[!t]
	\caption{Description of class label along with distribution of each class (in \%) in the acquired dataset}\label{tab:datasetCharacteristics}
	\centering
	\setlength{\tabcolsep}{5pt}
	\renewcommand{\arraystretch}{1.25}
	\begin{tabular}{p{0.28\linewidth}p{0.460\linewidth}c}
		\toprule
		{\bfseries Class label} &  {\bfseries Description} & {\bfseries Class\%} \\
		\toprule
		Appreciation & Citizen provided appreciative feedback. & $43.1$\% \\
		Satisfied & Citizen satisfied with the service. & $31.1$\% \\
		Peripheral complaint & Complains about peripheral service like non-availability of parking or complexity of the procedure. & $8.2$\% \\
		Demanded inquiry & More inquiry is required on the complaint. & $5.7$\% \\
		Corruption & Citizen reported bribery. & $3.5$\% \\
		Lagged response & Department responded with delay. & $2.1$\% \\
		Unresponsive & No response received by the citizen from the department. & $2.0$\% \\
		Medicine payment & Complainant was asked to buy basic medicine on his expense. & $1.8$\% \\
		Adverse behavior & Aggressive/intolerant behavior of the staff towards the citizen. & $1.5$\% \\
		Resource nonexistence & Department lacks necessary resources. & $0.6$\% \\
		Grievance ascribed & Malfeasance/Abuse of powers/official misconduct/sexual harassment to the complainant. & $0.3$\% \\
		Obnoxious/irrelevant & The SMS was irrelevant to public services. & $0.2$\% \\
		
		\bottomrule
	\end{tabular}
\end{table}

The dataset consists of SMS feedbacks of the citizens of Pakistan on different public services availed by them. The objective of collecting these responses is to measure the performance of government departments rendering different public services. Preprocessing of the data is kept minimal. All records having only single word in SMS were removed as cleaning step. To construct the ``gold standard", $313,813$ samples are manually annotated into $12$ predefined categories by two annotators in supervision of a domain-expert. Involvement of the domain-expert was to ensure the practicality and quality of the ``gold standard". Finally, stratified sampling method was opted for splitting the data into train and test partitions with $80-20$ ratio (i.e., $80\%$ records for training and $20\%$ records for testing). This way, training split has $251,050$ records while testing split has $62,763$ records. The rationale behind stratified sampling was to maintain the ratio of every class in both splits. The preprocessed and annotated data along with train and test split is made available \footnote{https://github.com/haroonshakeel/bilingual\_sms\_classification}. Note that the department names and service availed by the citizens is mapped to an integer identifier for anonymity.

Class label ratios,  corresponding labels, and it's description are presented in Table~\ref{tab:datasetCharacteristics}.

 \begin{figure}[!tp]
	\centering
	\includegraphics[page=1, scale=0.425, trim={0 0 0 0},clip]{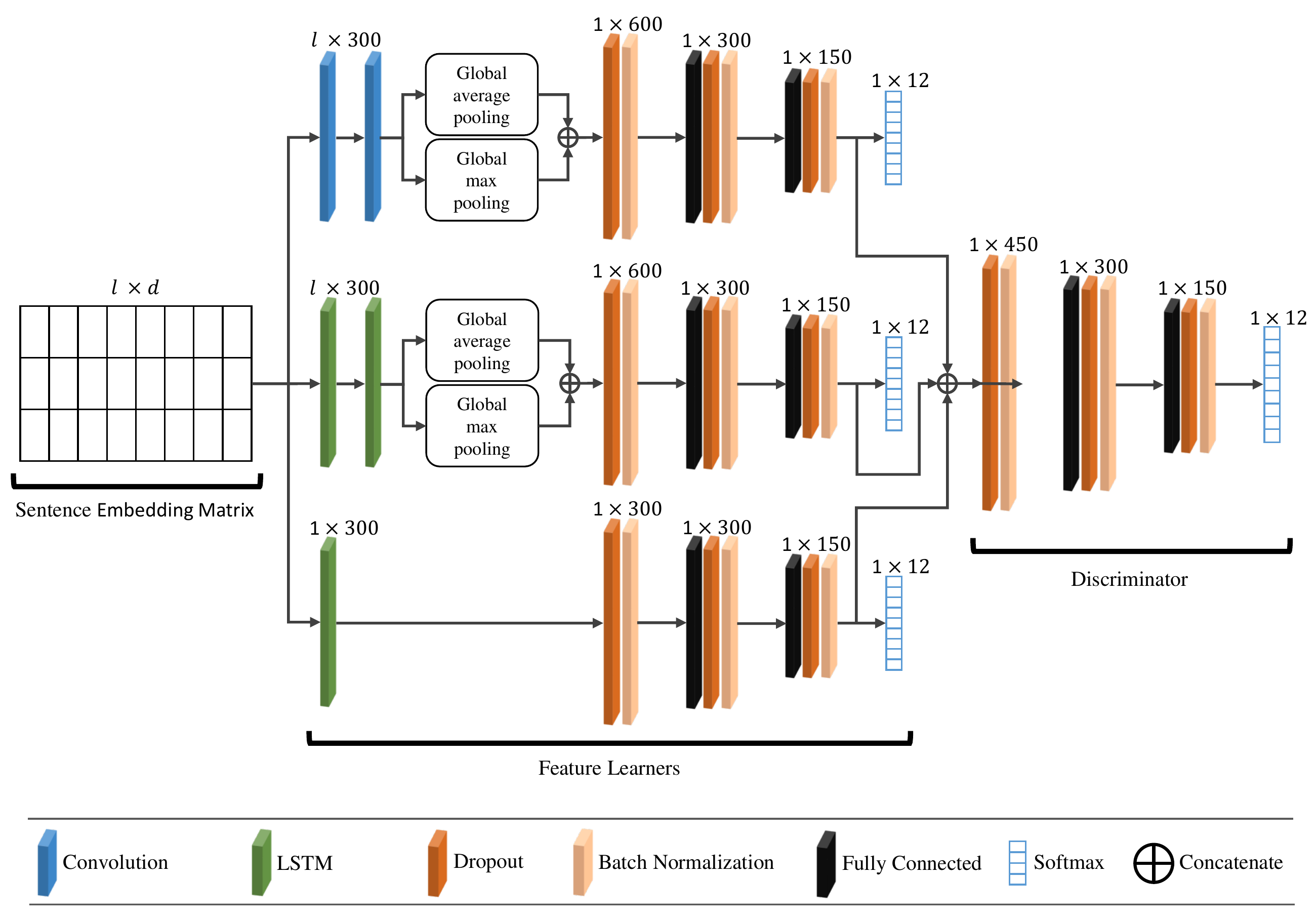}
	\caption{Multi-cascaded model (McM) for bilingual short text classification (figure best seen in color)}
	\label{fig:architecture}
\end{figure}

\section{Proposed Model and Experimentation} \label{sec:proposedModel}
 The proposed model, named \emph{McM}, is mainly inspired by the findings by Reimers, N., \& Gurevych (2017) , who concluded that deeper model have minimal effect on the predictive performance of the model~\cite{reimers2017reporting}. McM manifests a wider model, which employ three feature learners (cascades) that are trained for classification independently (in parallel). 
 
 The input text is first mapped to embedding matrix of size $l \times d$ where $l$ denotes the number of words in the text while $d$ is dimensions of the embedding vector for each of these words. More formally, let $\mathcal{T} \in \{w_1, w_2, ..., w_l\}$ be the input text with $l$ words, embedding matrix is defined by ${X} \in \mathbb{R}^{l \times d}$. This representation is then fed to three feature learners, which are trained with local supervision. The learned features are then forwarded to discriminator network for final prediction as shown in Fig.~\ref{fig:architecture}. Each of these components are discussed in subsequent subsections.

\subsection{Stacked-CNN Learner} \label{CNN_learner}
CNN learner is employed to learn $n$-gram features for identification of relationships between words. A $1$-d convolution filter is used with a sliding window (kernel) of size $k$ (number of $n$-grams) in order to extract the features. A filter $W$ is defined as $W \in \mathbb{R}^{k \times d}$ for the convolution function. The word vectors starting from the position $j$ to the position $j + k -1$ are processed by the filter $W$ at a time. The window $h_j$ is expressed as:
	
	\begin{equation} \label{eq:cnn_window}
		h_j = [X_j \oplus X_{j+1} \oplus, ..., \oplus X_{j+k-1}]
	\end{equation}
Where, the $\oplus$ represents the concatenation of word vectors. The number of filters are usually decided empirically. Each filter convolves with one window at a time to generate a feature map $f_j$ for that specific window as:
	
	\begin{equation}\label{eq:convolution}
		f_j = \sigma(h_j \odot W + b)
	\end{equation}
Where, the $\odot$ represents convolution operation, $b$ is a bias term, and $\sigma$ is a nonlinear transformation function \textit{ReLU}, which is defined as $\sigma(x) = max(x,0)$. The feature maps of each window are concatenated across all filters to get a high level vector representation and fed as input to next CNN layer. Output of second CNN layer is followed by (i) global max-pooling to remove low activation information from feature maps of all filters, and (ii) global average-pooling to get average activation across all the $n$-grams.

 These two outputs are then concatenated and forwarded to a small feedforward network having two fully-connected layers, followed by a \textit{softmax} layer for prediction of this particular learner. Dropout and batch-normalization layers are repeatedly used between both fully-connected layers to avoid features co-adaptation~\cite{srivastava2014dropout,ioffe2015batch}.
 
\subsection{Stacked-LSTM Learner} \label{subsec:stacked-lstm_learner}
The traditional methods in deep learning do not account for previous information while processing current input. LSTM, however, is able to memorize past information and correlate it with current information~\cite{wang-etal-2016-combination}. LSTM structure has memory cells (aka LSTM cells) that store the information selectively. Each word is treated as one time step and is fed to LSTM in a sequential manner. While processing the input at current time step $X_t$, LSTM also takes into account the previous hidden state $h_{t-1}$. The LSTM represents each time step with an input, a memory, and an output gate, denoted as $i_t, f_t$ and $o_t$ respectively. The hidden state $h_t$ of input $X_t$ for each time step $t$ is given by:

\begin{equation}
i_t = \sigma(W_iX_t + U_ih_{t-1} + b_i),
\end{equation}
\begin{equation}
f_t = \sigma(W_fX_t + U_fh_{t-1} + b_f),
\end{equation}
\begin{equation}
o_t = \sigma(W_oX_t + U_oh_{t-1} + b_o),
\end{equation}
\begin{equation}
u_t = tanh(W_u + U_uh_{t-1} + b_u),
\end{equation}
\begin{equation}
c_t = i_t * u_t + f_t * c_{t-1},
\end{equation}
\begin{equation}
h_t = o_t * tanh(c_t).
\end{equation}
Where, the $*$ is element-wise multiplication and $\sigma$ is sigmoid activation function.

Stacked-LSTM learner is comprised of two LSTM layers. Let ${H_1}$ be a matrix consisting of output vectors $\{h_1, h_2, ..., h_l\}$ that the first LSTM layer produced, denoting output at each time steps. This matrix is fed to second LSTM layer. Similarly, second layer produces another output matrix $H_2$ which is used to apply global max-pooling and global-average pooling. These two outputs are concatenated and forwarded to a two layered feedforward network for intermediate supervision (prediction), identical to previously described stacked-CNN learner.
	
\subsection{LSTM Learner} \label{subsec:lstm_learner}
LSTM learner is employed to learn long-term dependencies of the text as described in~\cite{wang-etal-2016-combination}. This learner encodes complete input text recursively. It takes one word vector at a time as input and outputs a single vector. The dimensions of the output vector are equal to the number of LSTM units deployed. This encoded text representation is then forwarded to a small feedforward network, identical to aforementioned two learners, for intermediate supervision in order to learn features. This learner differs from stacked-LSTM learner as it learns sentence features, and not average and max features of all time steps (input words).

\subsection{Discriminator Network} \label{subsec:discriminator}
The objective of discriminator network is to aggregate features learned by each of above described three learners and squash them into a small network for final prediction. The discriminator employs two fully-connected layers with batch-normalization and dropout layer along with \textit{ReLU} activation function for non-linearity. The \textit{softmax} activation function with categorical cross-entropy loss is used on the final prediction layer to get probabilities of each class. The class label is assigned based on maximum probability. This is treated as final prediction of the proposed model. The complete architecture, along with dimensions of each output is shown in Fig.~\ref{fig:architecture}.

\subsection{Experimental Setup} \label{subsec:experimental_setup}

 Pre-trained word embeddings on massive data, such as GloVe~\cite{pennington-etal-2014-glove}, give boost to predictive performance for multi-class classification~\cite{subramani2019deep}. However, such embeddings are limited to English language only with no equivalence for Roman Urdu. Therefore, in this study, we avoid using any \emph{word-based} pre-trained embeddings to give equal treatment to words of each language. We perform three kinds of experiments. (1) Embedding matrix is constructed using ELMo embeddings~\cite{peters-etal-2018-deep}, which utilizes characters to form word vectors and produces a word vector with $d = 1024$. We call this variation of the model McM$_\textsubscript{E}$. (2) Embedding matrix is initialized randomly for each word with word vector of size $d = 300$. We refer this particular model as McM$_\textsubscript{R}$. (3) We train domain specific embeddings\footnote{These embeddings are also made available along with dataset. } using word2vec with word vector of size $d = 300$ as suggested in original study~\cite{mikolov2013distributed}. We refer to this particular model as McM$_\textsubscript{D}$. 
 
  Furthermore, we also introduce soft-attention~\cite{wang-etal-2016-bilingual} between two layers of CNN and LSTM (in respective feature learner) to evaluate effect of attention on bilingual text classification. Attention mechanism ``highlights" (assigns more weight) a particular word that contributes more towards correct classification. We refer to attention based experiments with subscript $A$ for all three embedding initializations. This way, a total of $6$ experiments are performed with different variations of the proposed model. To mitigate effect of random initialization of network weights, we fix the random seed across all experiments. We train each model for $20$ epochs and create a checkpoint at epoch with best predictive performance on test split.
 
 We re-implement the model proposed in~\cite{medrouk2017deep}, and use it as a baseline for our problem. The rationale behind choosing this particular model as a baseline is it's proven good predictive performance on multilingual text classification. For McM, the choices of number of convolutional filters, number of hidden units in first dense layer, number of hidden units in second dense layer,  and recurrent units for LSTM are made empirically. Rest of the hyperparameters were selected by performing grid search using $20\%$ stratified validation set from training set on McM$_\textsubscript{R}$. Available choices and final selected parameters are mentioned in Table~\ref{tab:hyperparameters}. These choices remained same for all experiments and the validation set was merged back into training set.
 
 \begin{table}[!htbp]
	\centering
	\setlength{\tabcolsep}{3.5pt}
	\renewcommand{\arraystretch}{1.1}
	\caption{Hyperparameter tuning, the selection range, and final choice}\label{tab:hyperparameters}
	\begin{tabular}{ccc}
		
		\toprule
		{\bfseries Hyperparameter} &  {\bfseries Possible Values} & {\bfseries Chosen Value} \\
		\toprule
		First CNN layer kernel size ($k$) &  1, 2, 3, 4, 5 & 1 \\
		Second CNN layer kernel size ($k$) &  1, 2, 3, 4, 5 & 2 \\
		Dropout rate  &  0.1, 0.2, 0.3, 0.4, 0.5 & 0.2 \\
		Optimizer &  Adam, Adadelta, SGD & Adam \\
		Learning rate &  0.001, 0.002, 0.003, 0.004, 0.005 & 0.002 \\
		
		\bottomrule
	\end{tabular}
\end{table}

\subsection{Evaluation Metrics} \label{subsec:evaluation_metrics}
We employed the standard metrics that are widely adapted in the literature for measuring multi-class classification performance. These metrics are \textit{accuracy, precision, recall,} and \textit{F1-score}, where latter three can be computed using micro-average or macro-average strategies~\cite{sokolova2009systematic}. In micro-average strategy, each instance holds equal weight and outcomes are aggregated across all classes to compute a particular metric. This essentially means that the outcome would be influenced by the frequent class, if class distribution is skewed. In macro-average however, metrics for each class are calculated separately and then averaged, irrespective of their class label occurrence ratio. This gives each class equal weight instead of each instance, consequently favoring the under-represented classes. 

In our particular dataset, it is more plausible to favor smaller classes (i.e., other than ``Appreciation" and ``Satisfied") to detect potential complaints. Therefore, we choose to report macro-average values for precision, recall, and F1-score which are defined by (\ref{eq:precision}), (\ref{eq:recall}), and (\ref{eq:f1}) respectively.
\begin{equation} \label{eq:precision}
	Precision = \frac{\sum_{i=1}^{C} \frac{TP_i}{TP_i + FP_i}}{C},
\end{equation}
\begin{equation} \label{eq:recall}
Recall = \frac{\sum_{i=1}^{C} \frac{TP_i}{TP_i + FN_i}}{C},
\end{equation}
\begin{equation} \label{eq:f1}
F1-score = \frac{\sum_{i=1}^{C} \frac{2 \times Precision_i \times Recall_i}{Precision_i + Recall_i}}{C}.
\end{equation}
\section{Results and Discussion} \label{sec:results}
Before evaluating the McM, we first tested the baseline model on our dataset. Table~\ref{tab:results} presents results of baseline and all variations of our experiments. We focus our discussion on F1-score as accuracy is often misleading for dataset with unbalanced class distribution. However, for completeness sake, all measures are reported. 

\begin{table}[!t]
	\centering
	\setlength{\tabcolsep}{3pt}
	\caption{Performance evaluation of variations of the proposed model and baseline. Showing highest scores in boldface.}\label{tab:results}
	\begin{tabular}{llcccc}
		
		\toprule
		{\bfseries Model} & {\bfseries Component} & {\bfseries Accuracy} & {\bfseries Precision} & {\bfseries Recall} & {\bfseries F1-score} \\
		\toprule
		Baseline~\cite{medrouk2017deep} & - & $0.68$ & $0.52$ & $0.37$ & $0.39$ \\
		\midrule
		McM$_\textsubscript{E}$ & Stacked-CNN Learner & ${0.83}$ & ${0.66}$ & ${0.62}$ & ${0.63}$ \\
		& Stacked-LSTM Learner &${0.84}$ & ${0.70}$ & ${0.60}$ & ${0.64}$ \\
		& LSTM Learner &${0.80}$ & ${0.69}$ & ${0.48}$ & ${0.51}$ \\
		& Discriminator &${0.84}$ & ${0.68}$ & ${0.63}$ & ${0.66}$ \\
		\midrule
		McM$_\textsubscript{EA}$& Stacked-CNN Learner &  $0.82$ & $0.65$ & $0.57$ & $0.60$ \\
		& Stacked-LSTM Learner &  $0.82$ & $0.65$ & $0.57$ & $0.60$ \\
		& LSTM Learner &  $0.80$ & $0.62$ & $0.49$ & $0.51$ \\
		& Discriminator &  $0.83$ & $0.66$ & $0.60$ & $0.62$ \\
		\midrule
		McM$_\textsubscript{R}$ & Stacked-CNN Learner & $0.82$ & $0.66$ & $0.59$ & $0.62$ \\
		& Stacked-LSTM Learner & $0.82$ & $0.66$ & $0.58$ & $0.61$ \\
		& LSTM Learner & $0.81$ & $0.62$ & $0.59$ & $0.59$ \\
		& Discriminator & $0.83$ & $0.64$ & $0.61$ & $0.62$ \\
		\midrule
		McM$_\textsubscript{RA}$ & Stacked-CNN Learner &$0.80$ & $0.65$ & $0.52$ & $0.53$\\
		& Stacked-LSTM Learner &$0.81$ & $0.65$ & $0.55$ & $0.58$\\
		& LSTM Learner &$0.81$ & $0.64$ & $0.55$ & $0.58$\\
		& Discriminator &$0.81$ & $0.64$ & $0.58$ & $0.59$\\
		\midrule
		McM$_\textsubscript{D}$ & Stacked-CNN Learner &$0.84$ & $0.71$ & $0.63$ & $0.66$\\
		& Stacked-LSTM Learner &$0.85$ & $0.71$ & $0.67$ & $0.69$\\
		& LSTM Learner &$0.83$ & $0.68$ & $0.60$ & $0.63$\\
		& Discriminator & $\textbf{0.86}$ & $\textbf{0.72}$ & $\textbf{0.68}$ & $\textbf{0.69}$\\
		\midrule
		McM$_\textsubscript{DA}$ & Stacked-CNN Learner &  $0.82$ & $0.66$ & $0.59$ & $0.62$\\
		& Stacked-LSTM Learner &  $0.84$ & $0.69$ & $0.64$ & $0.66$\\
		& LSTM Learner &  $0.83$ & $0.67$ & $0.61$ & $0.63$\\
		& Discriminator &  $0.85$ & $0.70$ & $0.66$ & $0.67$\\
		\bottomrule
	\end{tabular}
\end{table}

It is observed from the results that baseline model performs worst among all the experiments. The reason behind  this degradation in performance can be traced back to the nature of the texts in the datasets (i.e., datasets used in original paper of baseline model~\cite{medrouk2017deep} and in our study). The approach in base model measure the performance of the model on multilingual dataset in which there is no code-switching involved. The complete text belongs to either one language or the other. However, in our case, the SMS text can have code-switching between two language, variation of spelling, or non-standard grammar. Baseline model is simple $1$ layered CNN model that is unable to tackle such challenges. On the other hand, McM learns the features from multiple perspectives, hence feature representations are richer, which consequently leads to a superior predictive performance. As every learner in McM is also supervised, all $4$ components of the proposed model (i.e., stacked-CNN learner, stacked-LSTM learner, LSTM-learner, and discriminator) can also be compared with each other.

In our experiments, the best performing variation of the proposed model is McM$_\textsubscript{D}$. On this particular setting, discriminator is able to achieve an F1-score of $0.69$ with precision and recall values of $0.72$ and $0.68$ respectively. Other components of McM also show the highest stats for all performance measures. However, for McM$_\textsubscript{DA}$, a significant reduction in performance is observed, although, attention-based models have been proven to show improvement in performance \cite{wang-etal-2016-bilingual}. Investigating the reason behind this drop in performance is beyond the scope of this study. The model variations trained on ELMo embedding have second highest performance. Discriminator of McM$_\textsubscript{E}$ achieves an F1-score of $0.66$, beating other learners in this experiment. However, reduction in performance is persistent when attention is used for McM$_\textsubscript{EA}$.

 \begin{figure}[!t]
 	\begin{tabular}{cc}
 		\centering
 		\includegraphics[width=.5\linewidth]{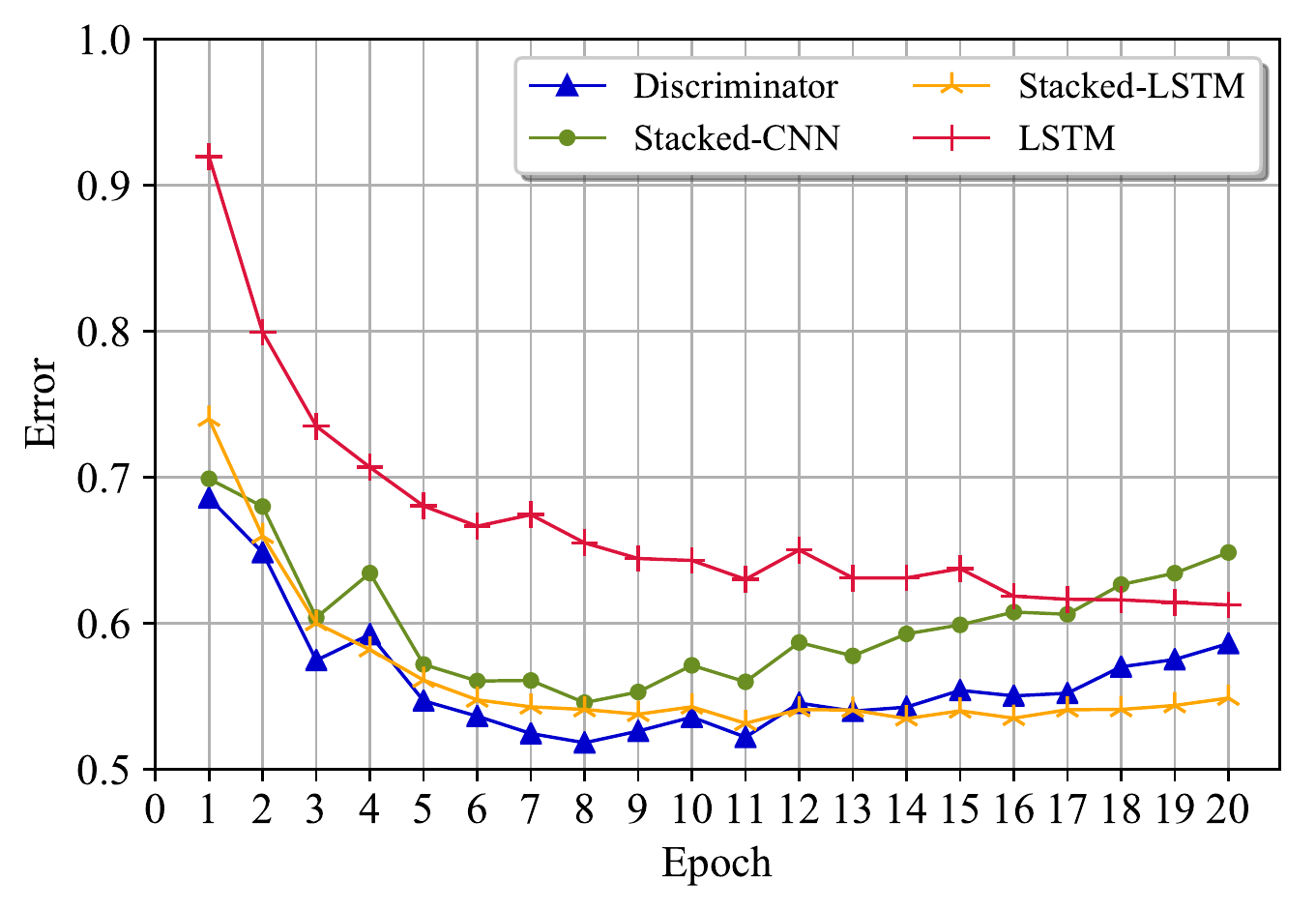}&
 		\includegraphics[width=.5\linewidth]{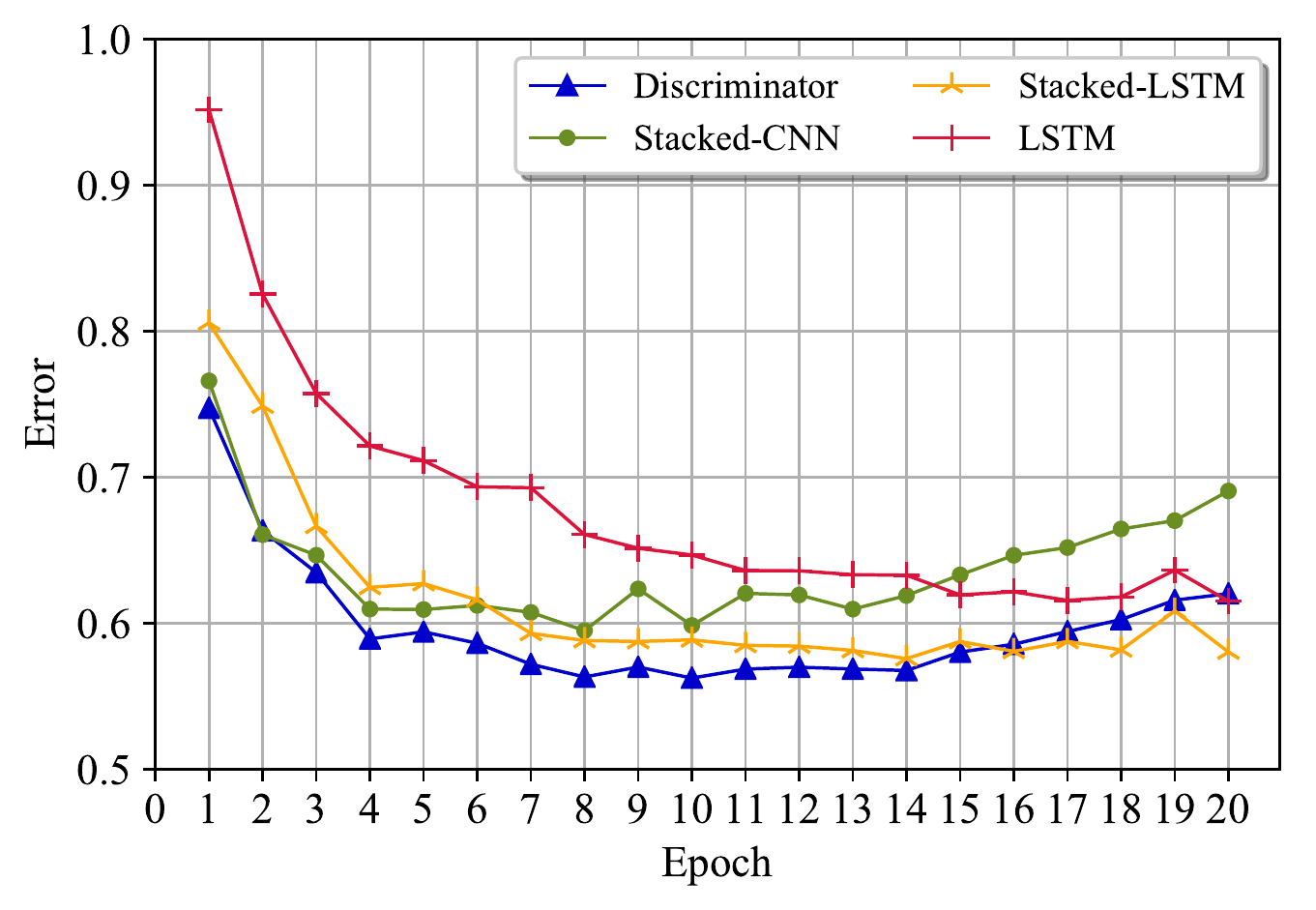}\\
 		\hspace{0.5cm}(a) McM$_\textsubscript{E}$  & \hspace{0.5cm}(b) McM$_\textsubscript{EA}$  \vspace{1.5mm}\\
 		\includegraphics[width=.5\linewidth]{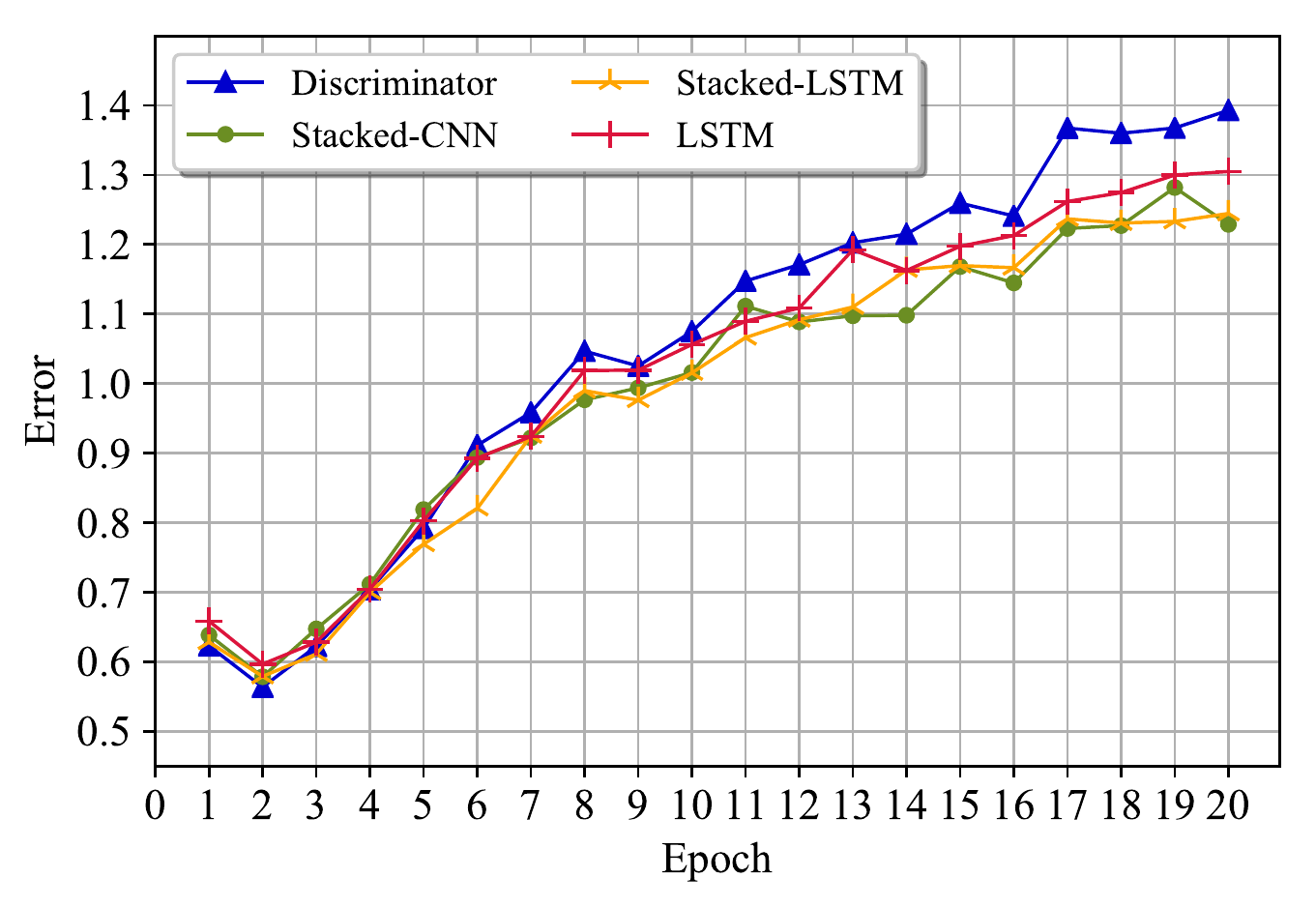}&
 		\includegraphics[width=.5\linewidth]{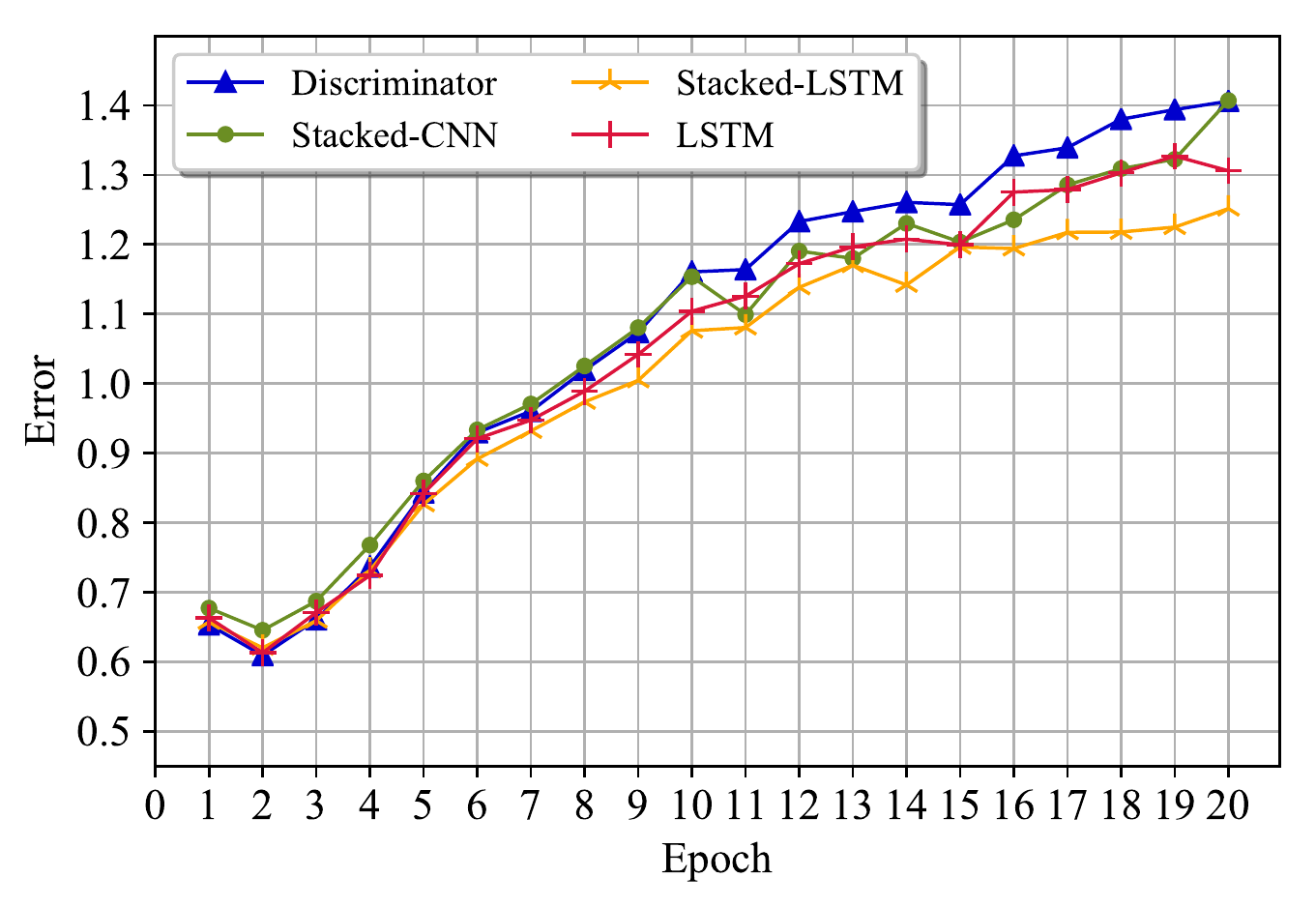}\\
 		\hspace{0.5cm}(c) McM$_\textsubscript{R}$   & \hspace{0.5cm}(d) McM$_\textsubscript{RA}$ \vspace{1.5mm}\\
 		\includegraphics[width=.5\linewidth]{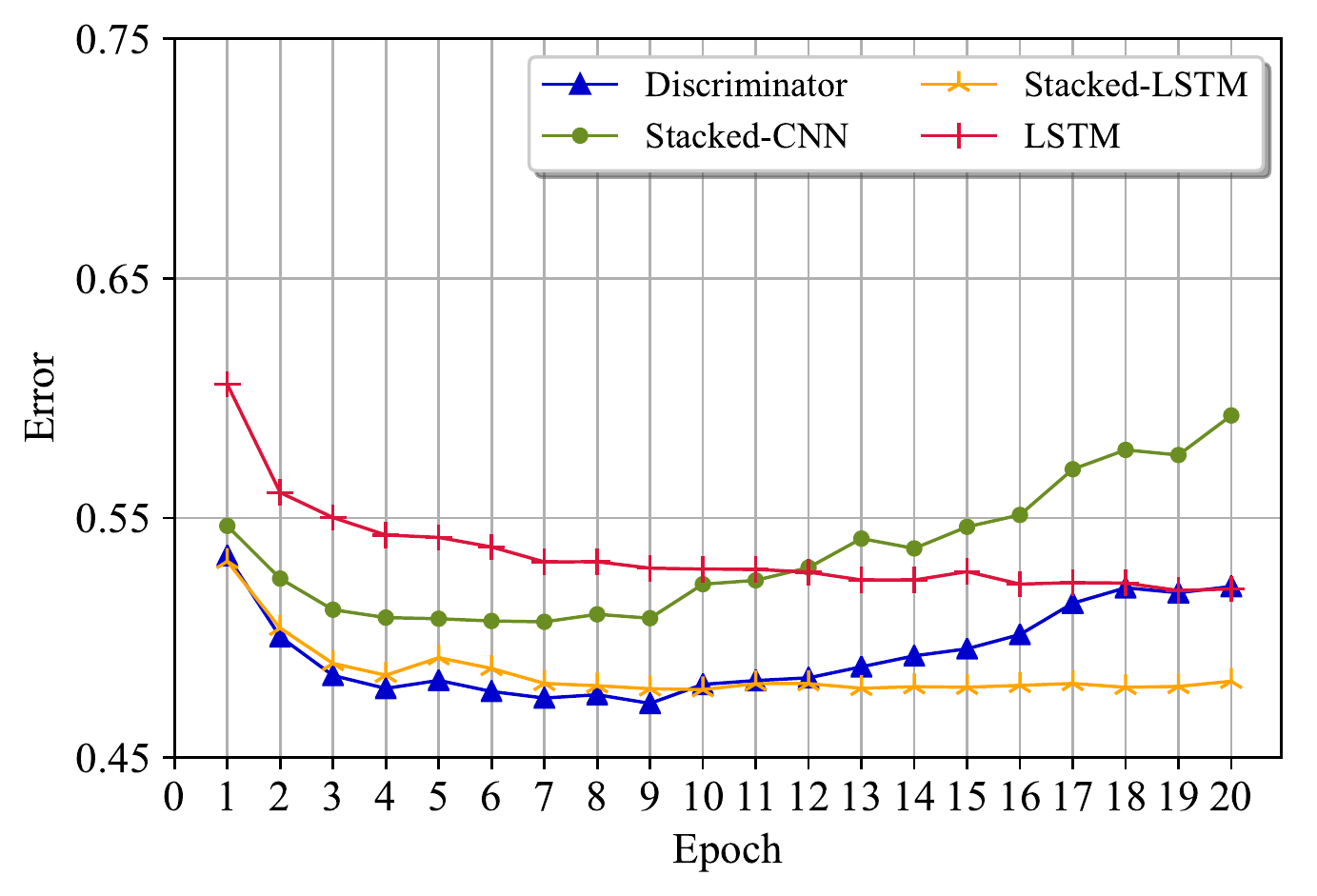}&
 		\includegraphics[width=.5\linewidth]{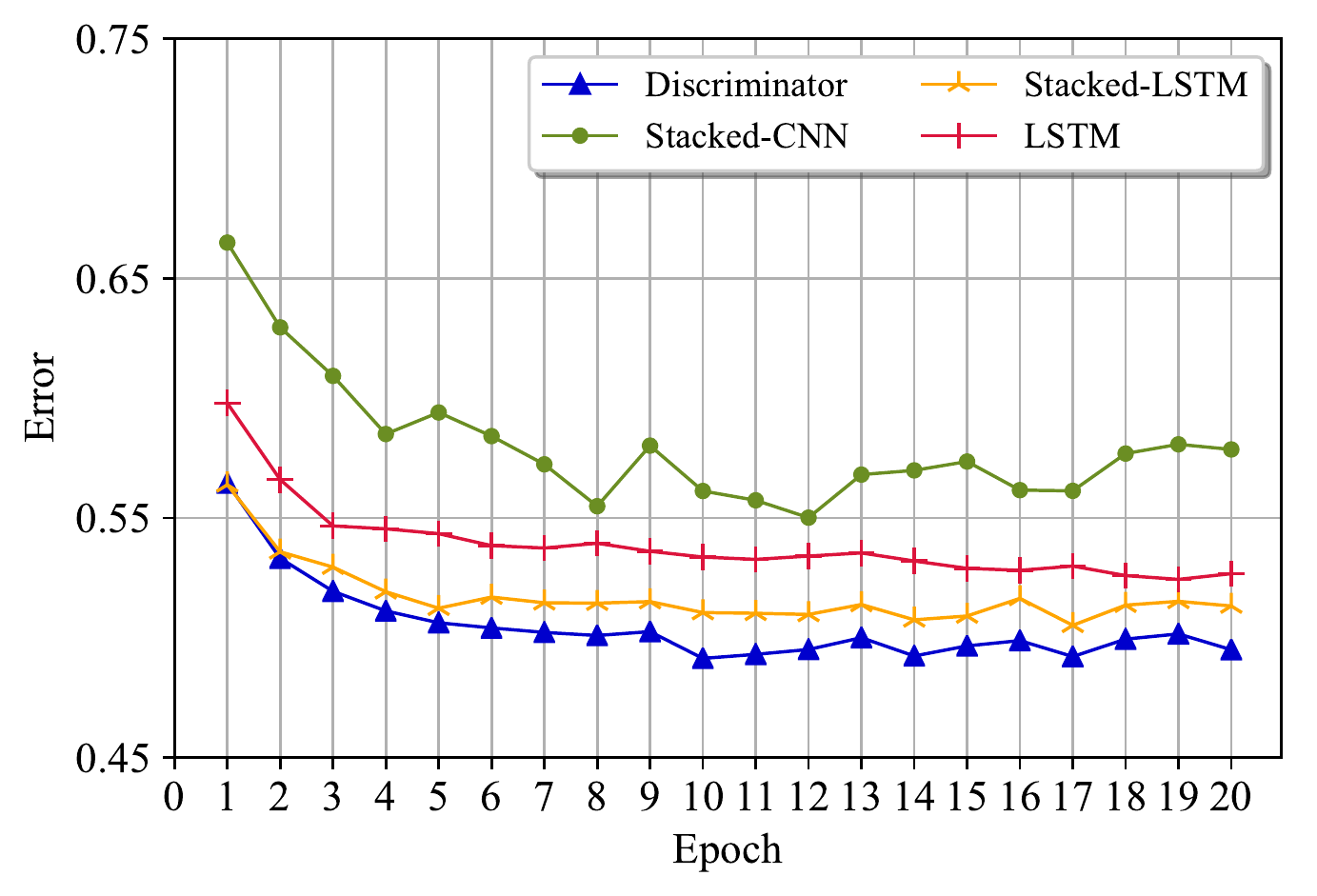}\\
 		\hspace{0.5cm}(e) McM$_\textsubscript{D}$   & \hspace{0.5cm}(f) McM$_\textsubscript{DA}$
 	\end{tabular}
 	\caption{Test error for all three feature learners and discriminator network over the epochs for all $4$ variations of the model, showing lowest error for domain specific embeddings while highest for random embedding initialization.}
 	\label{fig:LossPlots}
 \end{figure}
Regarding the experiments with random embedding initialization, McM$_\textsubscript{R}$ shows similar performance to McM$_\textsubscript{EA}$, while McM$_\textsubscript{RA}$ performs the worst. It is worth noting that in each experiment, discriminator network stays on top or performs equally as compared to other components in terms of F1-score. This is indication that discriminator network is able to learn richer representations of text as compared to methods where only single feature learner is deployed.
 
 Furthermore, the results for testing error for each component (i.e., $3$ learners and a discriminator network) for all $4$ variations of the proposed model are presented in Fig.~\ref{fig:LossPlots}. It is evident that the least error across all components is achieved by McM$_\textsubscript{D}$ model. Turning now to individual component performance, in ELMo embeddings based two models, lowest error is achieved by discriminator network, closely followed by stacked LSTM learner and stacked-CNN learner, while LSTM learner has the highest error. As far as model variations with random embeddings initializations are concerned, most interesting results are observed. As shown in subplot (c) and (d) in Fig.~\ref{fig:LossPlots}, McM$_\textsubscript{R}$ and McM$_\textsubscript{RA}$ tend to overfit. After second epoch, the error rate for all components of these two variations tend to increase drastically. However, it shows minimum error for discriminator in both variations, again proving that the features learned through multiple cascades are more robust and hold greater discriminative power. Note that in all $6$ variations of experiments, the error of discriminator network is the lowest as compared to other components of McM. Hence it can be deduced that learning features through multiple perspectives and aggregating them for final prediction is more fruitful as compared to single method of learning.

\section{Concluding Remarks} \label{sec:conclusion}
In this work, a new large-scale dataset and a novel deep learning architecture for multi-class classification of bilingual (English-Roman Urdu) text with code-switching is presented. The dataset is intended for enhancement of petty corruption detection in public offices and provides grounds for future research in this direction. While deep learning architecture is proposed for multi-class classification of bilingual SMS without utilizing any external resource. Three word embedding initialization techniques and soft-attention mechanism is also investigated. The observations from extensive experimentation led us to conclude that: (1) word embeddings vectors generated through characters tend to favor bilingual text classification as compared to random embedding initialization, (2) the attention mechanism tend to decrease the predictive performance of the model, irrespective of embedding types used, (3) using features learned through single perspective yield poor performance for bilingual text with code-switching, (4) training domain specific embeddings on a large corpus and using them to train the model achieves the highest performance.

With regards to future work, we intend to investigate the reason behind degradation of model performance with soft-attention.

%
%
%
\bibliographystyle{splncs04}
\bibliography{mybibliography}

\end{document}